%% file: main.tex
\documentclass[11pt]{article}
\usepackage[preprint]{acl}
\usepackage{times,latexsym}
\usepackage[T1]{fontenc}
\usepackage[utf8]{inputenc}
\usepackage{microtype,inconsolata,graphicx,booktabs,amsmath,amssymb,xcolor}
\usepackage{placeins,adjustbox,algorithm,algpseudocode}
\newcommand{\apbest}[1]{{\bfseries\boldmath #1}}
\newcommand{\apsecond}[1]{\underline{#1}}

\newcommand{\appending}[1]{\textcolor{gray}{#1}}
\newcommand{\apnote}[1]{}

\usepackage[normalem]{ulem}
\usepackage{colortbl}
\newcommand{\lcstyle}{\setlength{\aboverulesep}{0pt}\setlength{\belowrulesep}{0pt}\setlength{\extrarowheight}{1.2pt}\renewcommand{\arraystretch}{1.12}}

\newcommand{\method}{RangeSWA}
\newcommand{\lite}{RangeSWA-Lite}

\input{generated/kernel_v2_macros.tex}

\input{generated/revision_scope.tex}
\input{generated/enhanced_swa_macros.tex}
\input{generated/cola_macros.tex}

\input{generated/headline_macros.tex}
\input{generated/results_macros.tex}
\title{RangeSWA: Training-Free Length Extrapolation\\for Continuous Diffusion Language Models}

\author{
Guangfu Guo\textsuperscript{1},
Xiaoqian Lu\textsuperscript{2},
Linsey Pang\textsuperscript{3},
Weiran Yao\textsuperscript{4}\\[0.35ex]
\textbf{Haolin Chen\textsuperscript{4},
Long Cheng\textsuperscript{1},
Kunpeng Liu\textsuperscript{1,*}}\\[0.9ex]
{\normalfont\small
\textsuperscript{1}\,Clemson University\quad
\textsuperscript{2}\,University of Ottawa\quad
\textsuperscript{3}\,PayPal\quad
\textsuperscript{4}\,actAVA AI}\\[0.6ex]
{\normalfont\footnotesize\texttt{gguo@clemson.edu, xlu053@uottawa.ca, panglinsey@gmail.com}}\\
{\normalfont\footnotesize\texttt{weiran@actava.ai, haolin@actava.ai, lcheng2@clemson.edu, kunpenl@clemson.edu}}\\[0.5ex]
{\normalfont\footnotesize\textsuperscript{*}\,Corresponding authors}
}
\begin{document}
\renewcommand{\topfraction}{.85}\renewcommand{\textfraction}{.1}
\maketitle
\begin{abstract}
Continuous diffusion language models show promising potential in text generation tasks, but are prone to problems such as text repetition, semantic drift, and generation quality degradation when the generation length exceeds the training length. To address these challenges, we propose \textbf{RangeSWA}, a training-free length extrapolation method. The method combines local sliding-window attention, remote block summaries, and dynamic content retrieval to preserve local details and support long-range context access. In addition, generation degradation caused by length extrapolation is mitigated through positional remapping, conditioning attention correction, and denoising-stage-adaptive window scheduling. We conducted experiments on multiple continuous diffusion language models, and the results show that RangeSWA can effectively reduce generative perplexity and text repetition rates in long-text generation.
\end{abstract}

\section{Introduction}
In recent years, continuous diffusion language models have emerged as a promising research direction in text generation \citep{li2022diffusionlm,gong2023diffuseq}. Unlike traditional autoregressive models that generate text token by token \citep{radford2019language}, continuous diffusion language models update the entire sequence through iterative denoising and leverage bidirectional contextual information for text generation \citep{hu2026elf,chen2026langflow,guo2026cola}. However, these models are typically trained on sequences of limited length. When the generation length exceeds the training length, models are prone to text repetition, semantic drift, and generation quality degradation, which limits their application to long-text generation tasks. Such degeneration can also be hidden by a low generative perplexity \citep{zhang2026repetition,franca2026hacking}.

\begin{figure}[t]
\centering\includegraphics[width=\columnwidth]{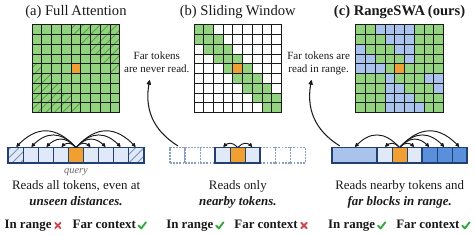}
\caption{\textbf{Context access of different attention mechanisms under length extrapolation.} (a) Full attention accesses all tokens but may exceed the trained position range. (b) Sliding-window attention attends only to nearby tokens and ignores distant information. (c) \method{} uses both local and distant information and maps distant positions back into the trained range.}
\label{fig:scenario}
\end{figure}

For length extrapolation in language models, prior work has proposed a variety of training-free methods, but these methods mainly target autoregressive (AR) language models. They typically extend the effective context length by adjusting positional encodings or modifying the attention mechanism. For example, positional interpolation (PI) \citep{chen2023extending} and positional scaling \citep{bloc972023ntk,peng2024yarn} adjust rotary position embeddings (RoPE) \citep{su2024roformer} to alleviate positional mismatches beyond the training length, while sliding-window attention (SWA) preserves local information by restricting the attention span \citep{beltagy2020longformer,xiao2024efficient,jolicoeur2026sliding}. In addition, Self-Extend \citep{jin2024llm} and LM-Infinite \citep{han2024lminfinite} also provide training-free context extension. However, because continuous diffusion language models rely on bidirectional iterative denoising, the applicability of existing length extrapolation methods and their effect on generation stability remain to be studied.

As shown in Figure~\ref{fig:scenario}, full attention encounters positional distances unseen during training, which may lead to generation degradation. In contrast, sliding-window attention avoids excessively long positional distances but loses access to distant contextual information, limiting cross-paragraph interactions. Moreover, continuous diffusion language models update the entire sequence at each denoising step, and different stages may require different amounts of contextual information. Further analysis shows that sliding-window attention may also alter the model's attention allocation to conditioning information. Even when some prefix tokens are retained, this effect cannot be completely eliminated. Therefore, existing methods still face three main challenges: (1) positional mismatches beyond the training range; (2) unstable conditioning attention; and (3) limited access to distant contextual information.

To address these challenges, we propose \method{}, a training-free length extrapolation method for continuous diffusion language models. The method combines local sliding-window attention, remote block summaries, and dynamic content retrieval into a unified attention mechanism to preserve local information while enabling long-range context access. Meanwhile, we introduce positional remapping and conditioning attention correction to alleviate positional and conditioning mismatches caused by length extrapolation. In addition, denoising-stage-adaptive window scheduling adjusts the attention span across different denoising stages, further improving the stability of long-text generation. The proposed method requires no updates to pretrained model parameters and can be directly applied to existing models.

We conduct length extrapolation experiments from 1k to 8k tokens on three Embedded Language Flows (ELF) models, ELF-B/M/L \citep{hu2026elf}, and further evaluate the applicability of the method to CoLa \citep{guo2026cola} and LangFlow \citep{chen2026langflow}. Experimental results show that, compared with standard sliding-window attention, \method{} effectively reduces generative perplexity and text repetition rates in long-text generation on ELF-M/L. Ablation experiments further demonstrate that conditioning attention correction and adaptive window scheduling account for most of the perplexity improvements, while remote context access further reduces text repetition.

We summarize our main contributions as follows:
\begin{enumerate}
\item We identify three key challenges in length extrapolation for continuous diffusion language models: positional mismatches, conditioning attention instability, and limited access to distant contextual information.
\item We propose \method{}, a training-free length extrapolation method that combines local and remote attention mechanisms, positional remapping, conditioning attention correction, and adaptive window scheduling to mitigate long-text generation degradation.
\item We conduct experiments on ELF-B/M/L, CoLa, and LangFlow to demonstrate the effectiveness of \method{} in reducing generative perplexity and text repetition rates, and analyze the contributions of individual components through ablation studies.
\end{enumerate}

\section{Related Work}
\label{sec:related}
\paragraph{Sparse and Retrieval Attention.} To reduce the cost of long-sequence modeling, existing methods use sliding windows, global tokens, attention sinks, and content-based retrieval \citep{beltagy2020longformer,xiao2024efficient,tang2024quest,yuan2025native}. LM-Infinite extends context by bounding positional distances \citep{han2024lminfinite}, while prefix-preserving sliding windows also support frozen-model extrapolation \citep{jolicoeur2026sliding}. Compared with autoregressive decoding, continuous diffusion repeatedly updates the entire sequence and uses bidirectional context at each denoising step. RangeSWA therefore combines remote block summaries with dynamic content retrieval to adapt context access to the evolving denoising states.

\paragraph{Length Extrapolation.} Context-extension methods adapt positional encodings through interpolation, frequency scaling, or grouped remapping, including PI, neural tangent kernel (NTK) scaling, YaRN, and Self-Extend \citep{chen2023extending,bloc972023ntk,peng2024yarn,jin2024llm}. These approaches mitigate positional distances beyond the training range. Their original recipes differ in whether they require fine-tuning; our baseline comparisons use frozen inference-time adaptations (Appendix~\ref{app:impl}). In our ELF experiments, positional changes alone do not prevent length-induced degeneration, and extrapolation also changes conditioning interactions. RangeSWA complements positional remapping with conditioning attention correction and distant-context access to mitigate this additional degeneration.

\paragraph{Diffusion Language Models.} Diffusion models have emerged as an alternative paradigm for text generation. Continuous models jointly update the entire sequence through iterative denoising \citep{hu2026elf,chen2026langflow}, whereas masked discrete diffusion updates masked tokens in parallel \citep{sahoo2024simple}. CoLa uses continuous latent variables for blockwise generation \citep{guo2026cola}. These architectures differ in their generation procedures, and extending their frozen checkpoints beyond the training length can introduce additional degeneration. We focus on continuous diffusion language models and propose RangeSWA to improve long-text generation stability without additional training.

\section{Methodology}
\label{sec:method}
\begin{figure*}[t]
\centering\includegraphics[width=\textwidth]{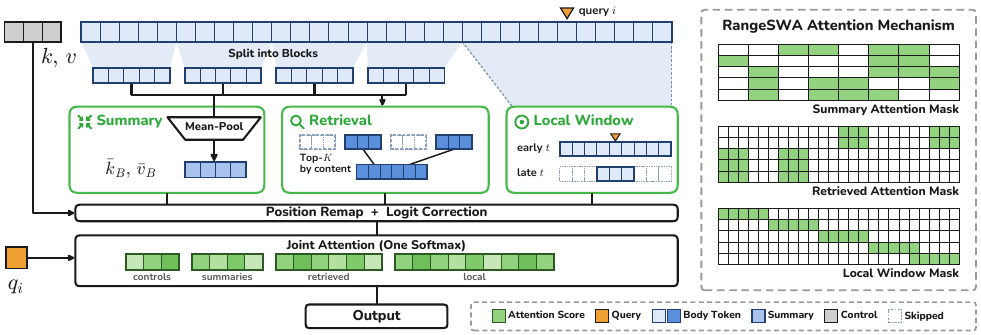}
\caption{\textbf{Overview of the \method{} attention mechanism.} Each body query attends to a local window, summaries of remote blocks and the top-$K$ retrieved blocks, together with control tokens, in a single softmax after positional remapping. The right panel shows the attention masks.}
\label{fig:method}
\end{figure*}

We propose \method{}, a training-free length extrapolation method that integrates local attention, remote block summaries, and dynamic content retrieval in one attention readout (Figure~\ref{fig:method}). Positional remapping, conditioning attention correction, and an adaptive window schedule mitigate degeneration at extended lengths. We first formulate the problem, then describe each component. The configuration is fixed on ELF-B and reused on M/L; architecture-specific adaptations are in Appendix~\ref{app:impl}.

\subsection{Problem Formulation and Overview}
\label{sec:prelim}
We describe whole-sequence denoising using ELF \citep{hu2026elf}. Let $T=1024$ be the training length and $L$ the target body length. At each denoising time $t$, every layer computes queries, keys, and values from the current representations. Body tokens use rotary position embeddings (RoPE) \citep{su2024roformer}; twelve unrotated control tokens encode time, self-conditioning guidance, and generation mode. Standard attention uses $a_{ij}=\operatorname{softmax}_j(q_i^\top k_j/\sqrt d)$, where $d$ is the head dimension and queries and keys include their positional transformations.

When $L>T$, full attention encounters unseen positional distances. SWA restricts the visible context and renormalizes attention over a retained set $\mathcal S_i$:
\begin{equation}
\widetilde a_{ij}=\frac{a_{ij}\mathbf{1}[j\in\mathcal S_i]}{\sum_{u\in\mathcal S_i}a_{iu}}.
\label{eq:renorm}
\end{equation}
This also changes the influence of retained conditioning tokens. \method{} jointly handles local context, distant information, and conditioning interactions without parameter updates.

\subsection{Local and Remote Attention}
\label{sec:access}
For body query $i$, the candidate set is $\mathcal A_i=\mathcal C\cup\mathcal N_i\cup\mathcal R_i\cup\mathcal B_i$: control tokens $\mathcal C$, a centered local window $\mathcal N_i$, retrieved exact tokens $\mathcal R_i$, and remote block summaries $\mathcal B_i$. All candidates share one softmax:
\begin{equation}
o_i=\sum_{c\in\mathcal A_i}\frac{\exp(\ell_{ic})}{\sum_{u\in\mathcal A_i}\exp(\ell_{iu})}\,v_c,
\label{eq:joint}
\end{equation}
where $\ell_{ic}$ and $v_c$ are the candidate score and value. The native output projection follows this readout.

\paragraph{Remote block summaries.} Body tokens are partitioned into blocks of $G=16$. Each eligible block $B$ is represented by the mean of its position-rotated keys $\overline k_B$ and its values $\overline v_B$. Its score is
\begin{equation}
\ell_{iB}=\frac{q_i^\top\overline k_B}{\sqrt d}+\log|B|.
\label{eq:summary}
\end{equation}
The count bias matches the mass of $|B|$ identical keys; for heterogeneous keys it is an approximation. Summaries using grouped positions receive an additional $-\log g$ correction. Blocks overlapping local or retrieved tokens are excluded (Appendix~\ref{app:impl}).

\paragraph{Dynamic content retrieval.} Queries are grouped into 128-token blocks $Q_b$, and keys into 64-token blocks $P_b$. Each attention head scores remote blocks using mean queries and keys before RoPE:
\begin{equation}
s(Q_b,P_b)=\overline q_{Q_b}^{\,\top}\overline k_{P_b}.
\label{eq:retrieval}
\end{equation}
Up to $K=8$ eligible blocks are selected per query block, and their original tokens enter $\mathcal R_i$. Overlapping summaries are removed. Selection is recomputed at every layer and model call to follow the evolving representations.

\subsection{Positional and Conditioning Correction}
\label{sec:poscond}
\paragraph{Positional remapping.} Local interactions and remote interactions within $H=512$ retain native RoPE positions. Farther interactions use Self-Extend's grouped coordinates \citep{jin2024llm}, with an effective query--key distance at most $T-1$. Retrieved tokens follow the same rules. Appendix~\ref{app:poscorr} summarizes the remapping. Positions remain unchanged when $L\le T$.

\paragraph{Conditioning attention correction.} Unrotated control tokens interact with body tokens through their absolute positions. For $L>T$, both control--body directions use the interpolated body coordinate $p_i^{\mathrm{cond}}=i(T-1)/(L-1)$; body--body interactions retain the rules above. Control queries retain global access to all control and body keys. Only their body-key scores receive the key-count correction
\begin{equation}
\ell_{cj}^{\mathrm{corr}}=\ell_{cj}-\log(L/T),\qquad L>T,
\label{eq:controlmass}
\end{equation}
where $c$ is a control query and $j$ a body key. This compensates for key-count growth; actual attention mass still depends on the scores. Both corrections are inactive when $L\le T$.

\subsection{Denoising-Stage-Adaptive Window}
\label{sec:schedule}
Context requirements may vary during denoising. \method{} uses a wider local window at high noise, a narrower window later, and an expanded window for token decoding:
\begin{equation}
W(t)=\begin{cases}
2047, & \text{decoding call},\\
1023, & t<0.182,\\
256, & t\ge0.182.
\end{cases}
\label{eq:window}
\end{equation}
Here $t=0$ denotes pure noise, and the decoding case takes precedence. The method replaces self-attention while retaining pretrained parameters, native projections, sampling, self-conditioning, and other network operations. Algorithm~\ref{alg:rangeswa} summarizes one attention call, and Appendix~\ref{app:impl} provides implementation and sampling settings.

\begin{algorithm}[t]
\caption{\method{} attention in one network call}
\label{alg:rangeswa}
\small
\begin{algorithmic}[1]
\Require body queries, keys and values; control keys; time $t$; length $L$
\State Set the local window $W(t)$ (Eq.~\ref{eq:window})
\State Split distant body keys into blocks and form summaries (Eq.~\ref{eq:summary})
\State Retrieve the top-$K$ distant blocks by content; they replace their summaries (Eq.~\ref{eq:retrieval})
\State Map far positions into the trained range (Section~\ref{sec:poscond}) and correct control interactions (Eq.~\ref{eq:controlmass})
\For{each body query}
  \State One softmax over controls, local window, retrieved tokens and summaries (Eq.~\ref{eq:joint})
\EndFor
\State Control queries attend globally with the mass correction
\State \Return the native output projection
\end{algorithmic}
\end{algorithm}

\begin{table*}[!t]
\centering\scriptsize\setlength{\tabcolsep}{4pt}
\resizebox{\textwidth}{!}{\input{generated/table1_elf.tex}}
\caption{\textbf{Length extrapolation on ELF-B/M/L at 1k--8k.} PPL, entropy $H$ and $R_4$(\%) over 1000 samples per configuration; markers are defined in Appendix~\ref{app:metrics}.}
\label{tab:elf}
\end{table*}

\subsection{Computational Cost and \lite{}}
\label{sec:cost}
\label{app:cost}
Each query requires computing scores for $W+|\mathcal R_i|$ exact keys, $2H/G$ near summaries, and approximately $(L-2H)/G$ far summaries. Retrieval also requires comparing $O(L^2/(128\cdot64))$ pairs of blocks. Therefore, the full \method{} still has quadratic terms, with a $1/G$ factor on the far-summary term. \lite{} removes the far field, retrieval and high-noise wide window, retaining summaries within $H=1024$, conditioning corrections and the bounded decoder window. It uses a fixed number of candidates per query, so the computational cost grows linearly with $L$.

\section{Experimental Setup}
\label{sec:setup}

\paragraph{Models and Settings.} We evaluate the official ELF-B/M/L checkpoints \citep{hu2026elf} at 1k, 2k, 4k, and 8k tokens, generating 1000 samples per configuration. Methods within each model and length share initial noise. We also evaluate transfer to CoLa \citep{guo2026cola} and LangFlow \citep{chen2026langflow}. Sampling and implementation details are in Appendix~\ref{app:impl}.

\paragraph{Baselines.} ELF baselines include Full Attention (Full), positional interpolation (PI) \citep{chen2023extending}, NTK scaling \citep{bloc972023ntk}, ELF-adapted Self-Extend* \citep{jin2024llm}, and sliding-window attention (SWA). All methods use frozen pretrained checkpoints without additional training. Baseline adaptations are in Appendix~\ref{app:impl}.

\paragraph{Evaluation Metrics.} We jointly report GPT-2 Large generative perplexity (Gen.PPL, reported as PPL in tables), mean document unigram entropy ($H$), and word four-gram repetition rate ($R_4$) as generation quality proxies. $H$ measures lexical diversity and $R_4$ measures repetition. We flag ELF configurations with mean $H<5.0$ to avoid low perplexity masking repetitive outputs. Lexical continuity is supplementary. Metric definitions are in Appendix~\ref{app:metrics}; comparisons use the same model and length.

\section{Results and Analysis}
\label{sec:results}

\subsection{Length Extrapolation Performance}
\label{sec:performance}
Table~\ref{tab:elf} reports the results of \method{} on ELF-B/M/L at different generation lengths. When the generation length exceeds the training length, full attention and positional adaptation methods usually show clear degradation in generation quality. In contrast, \method{} effectively reduces generative perplexity and text repetition on ELF-M/L, outperforming SWA, especially at longer generation lengths. On ELF-B, \method{} lowers perplexity but increases repetition, indicating that its effect differs across models.

\begin{figure}[t]
\centering\includegraphics[width=\columnwidth]{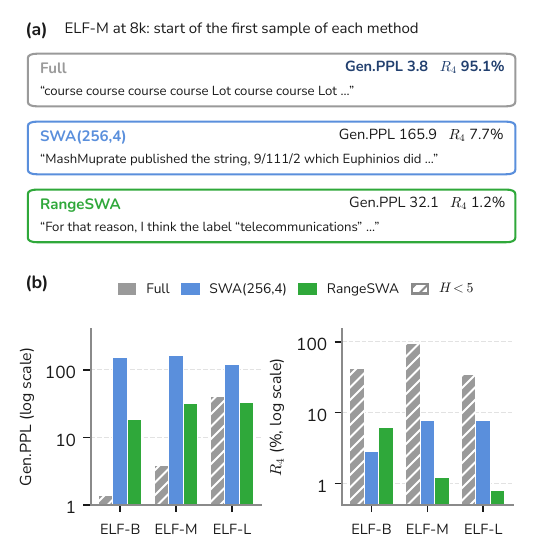}
\caption{\textbf{Low Gen.PPL can mask severe text repetition.} (a) Opening text of one sample per method on ELF-M at 8k. (b) Gen.PPL and $R_4$ on ELF-B/M/L at 8k; hatched bars mark $H<5$.}
\label{fig:extrapolation}
\end{figure}

Figure~\ref{fig:extrapolation} further shows that lower generative perplexity does not necessarily mean better generation quality, since severely repetitive text can also have low perplexity. Generation quality should therefore be assessed jointly with perplexity, lexical diversity and repetition \citep{franca2026hacking,zhang2026repetition}. Trends across generation lengths are shown in Figure~\ref{fig:trajectory} (Appendix~\ref{app:diag}).

\input{generated/table_seeds_all.tex}

To verify the stability of the results, we repeat the key experiments with three different random seeds. As shown in Table~\ref{tab:seedsall}, \method{} consistently outperforms SWA on ELF-M/L with small variation across seeds, further indicating the stability of the method for length extrapolation.

\subsection{Transfer to Other Backbones}
\label{sec:transfer}

To evaluate the generalization of \method{}, we further apply it to CoLa \citep{guo2026cola} and LangFlow \citep{chen2026langflow} while keeping their original generation procedures unchanged. The results are shown in Table~\ref{tab:transfer}. On CoLa, \method{} substantially reduces text repetition compared with SWA, with a repetition rate close to full attention. On LangFlow, \method{} performs close to full attention at the training length. As the generation length increases, SWA suffers from severe text repetition, whereas \method{} effectively mitigates it; for example, at 2k, \method{} reduces the repetition rate from 49.5\% to 5.3\%. Overall, \method{} effectively mitigates long-text generation degradation across different continuous diffusion language models, indicating a degree of generalization, although the size of the improvement varies across models.

\begin{table*}[t]
\centering\small\setlength{\tabcolsep}{5pt}
\input{generated/table_transfer_main.tex}
\caption{\textbf{Transfer to CoLa and LangFlow.} PPL, $H$ and $R_4$(\%); CoLa is scored conditionally after a 1k prompt and LangFlow unconditionally (Appendix~\ref{app:backbones}).}
\label{tab:transfer}
\label{tab:cola}
\label{tab:langflow}
\end{table*}

\subsection{Mechanism Analysis and Ablation Study}
\label{sec:ablation}
\begin{figure}[t]
\centering\includegraphics[width=\columnwidth]{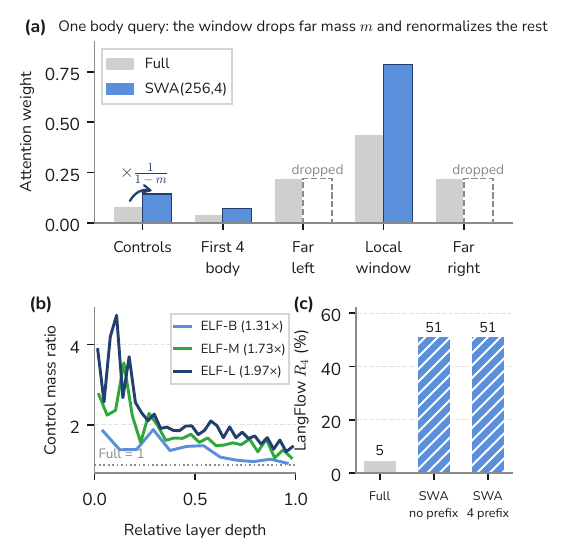}
\caption{\textbf{Retaining body prefixes is not enough for sliding-window attention.} (a) Dropping distant attention mass $m$ scales control-token weights by $1/(1-m)$. (b) Control-token mass under SWA(256,4) relative to Full, per layer at 1k. (c) With a 256-token window, LangFlow repeats heavily with or without prefixes.}
\label{fig:truncation}
\end{figure}
\begin{table}[t]
\centering\small\setlength{\tabcolsep}{4pt}
\begin{adjustbox}{max width=\columnwidth}\input{generated/table_ablation_main.tex}\end{adjustbox}
\caption{\textbf{ELF-B component deletions at 8k.} Each row removes one component from \method{}; 1000 outputs, seed 1001.}
\label{tab:ablationmain}
\label{tab:mechanism}
\end{table}

\begin{table}[t]
\centering\small\setlength{\tabcolsep}{3pt}
\begin{adjustbox}{max width=\columnwidth}\input{generated/table_ablation_remote.tex}\end{adjustbox}
\caption{\textbf{Simplified variants of \method{} at 8k.} Gen.PPL and $R_4$(\%) without remote access. $^\dagger$: simplified variant; C: conditioning corrections; S: window schedule.}
\label{tab:remote}
\end{table}

\begin{table}[t]
\centering\setlength{\tabcolsep}{3pt}\resizebox{\columnwidth}{!}{\input{generated/table_efficiency_main.tex}}
\caption{\textbf{bfloat16 (BF16) sampling latency on ELF-B.} Median time over five runs and paired speedup over Full on one H100 with batch size 16.}
\label{tab:efficiency}
\end{table}

\paragraph{Attention analysis.} To further analyze the effectiveness of \method{}, we analyze its attention mechanism and conduct ablation studies on its key components. We first examine how sliding-window attention affects the model. As shown in Figure~\ref{fig:truncation}(a), sliding-window attention discards the attention weights of distant tokens and renormalizes the remaining weights, which changes the model's original attention allocation and gives control tokens a higher relative attention weight. Retaining some tokens at the beginning of the sequence does not fully eliminate this effect, as shown in Figure~\ref{fig:truncation}(b).

Figure~\ref{fig:truncation}(c) further shows that on LangFlow, which has no control tokens, retaining only prefix tokens still cannot prevent severe text repetition. This suggests that generation degradation is related not only to changes in attention allocation but also to the loss of distant context. Moreover, although adding remote block summaries brings perplexity close to full attention, the lexical similarity between distant text segments remains low, indicating that coarse summaries alone cannot fully recover distant context.

\paragraph{Component ablations.} To verify the role of each component, we further conduct ablation experiments. As shown in Table~\ref{tab:ablationmain}, the conditioning attention corrections and the adaptive window schedule strongly affect generation quality. Removing the conditioning positional interpolation or restoring full attention in the final decoding stage both clearly increase perplexity: at ELF-B 8k, perplexity rises from 18.37 to 76.68 and 80.43, respectively, and a similar increase is observed on ELF-M/L when the interpolation is removed. Remote block summaries and dynamic retrieval also benefit generation quality. On ELF-B, removing either component increases perplexity. In addition, removing dynamic retrieval lowers the long-range lexical continuity of the text, indicating that dynamic retrieval helps maintain connections between different text segments. The positional mapping also affects generation. Restoring the original positions of distant tokens lowers perplexity but increases text repetition, indicating a trade-off between perplexity and repetition.

\paragraph{Simplified variants and trade-offs.} To further separate the contributions of different components, we design simplified variants of \method{}, shown in Table~\ref{tab:remote}. Even without distant context, adding only the conditioning attention corrections and the adaptive window schedule to a plain sliding window substantially reduces perplexity, indicating that these two mechanisms are important factors in improving long-text generation quality. In contrast, merely enlarging the sliding window does not effectively mitigate generation degradation. Even when the window is enlarged to an attention cost comparable to \method{} (B0; Appendix~\ref{app:ladderB}), perplexity remains high, whereas adding the conditioning attention corrections and a bounded decoding window clearly lowers it, further showing that the quality gain does not simply come from a larger attention range.

On the other hand, remote block summaries and dynamic retrieval further reduce text repetition, especially on ELF-M/L. Although the simplified variant without distant context sometimes reaches lower perplexity, the complete \method{} has lower text repetition. Removing dynamic retrieval clearly increases repetition on both models, showing that access to distant information plays an important role in reducing repeated content. Overall, the components of \method{} are complementary: the conditioning attention corrections and the adaptive window schedule mainly improve perplexity, while remote block summaries and dynamic retrieval further reduce text repetition. Nevertheless, different models still show a trade-off between perplexity and repetition. In addition, all hyperparameters were first determined on ELF-B and then applied directly to the other models. The sensitivity experiments show that different parameter settings also affect generation; for example, retrieving more blocks lowers perplexity but may increase repetition. Full results are in Appendices~\ref{app:revision}--\ref{app:enhancedswa}.

\subsection{Efficiency Analysis}
\label{sec:efficiency}
To evaluate the inference efficiency of \method{}, we measure the sampling latency of ELF-B at different generation lengths on a single H100 graphics processing unit (GPU); the results are shown in Table~\ref{tab:efficiency}. At shorter generation lengths, \method{} runs at roughly the same speed as full attention or slightly slower. As the generation length increases, its efficiency advantage gradually emerges, reaching a \SpRangeBFEight{} speedup over full attention at 8k.

\lite{} further reduces the computational cost by simplifying the remote attention computation, achieving a \SpLiteBFEight{} speedup at 8k, but its perplexity and text repetition also increase, indicating a trade-off between computational efficiency and generation quality.

Further analysis shows that \method{} effectively reduces the attention time, but the computation in the rest of the network limits the overall speedup (Table~\ref{tab:kernelprofile}). Meanwhile, the Triton implementation is numerically consistent with the reference implementation and leaves generation quality essentially unchanged. Overall, \method{} improves inference efficiency on longer sequences, but the full method still contains quadratic computational terms. In contrast, \lite{} is more efficient but trades generation quality for inference speed. Detailed results are in Appendix~\ref{app:eff}.

\begin{figure}[t]
\centering\scriptsize\input{generated/case_main.tex}
\caption{\textbf{A 2k example on ELF-B.} Opening and closing 16 words of sample 350 (seed 1001), with metrics on the full sample. The methods share only the seed and initial noise, so their topics differ.}
\label{fig:case}
\end{figure}

\subsection{Model-Based and Qualitative Evaluation}
\label{sec:modeleval}

To further evaluate long-text generation quality, we use Qwen3-14B to automatically evaluate the generated texts and complement it with a qualitative analysis of generation examples. Each comparison is repeated with the presentation order swapped to reduce order bias (Appendix~\ref{app:evaluation}). As shown in Table~\ref{tab:llmjudge}, the judge generally prefers text generated by \method{} over text generated by full attention and SWA, and the advantage becomes more pronounced as the generation length increases.

Figure~\ref{fig:case} further shows generation examples under different attention mechanisms. Text generated with full attention contains a large amount of repeated content, whereas SWA tends to drift away from the topic. In contrast, \method{} maintains a relatively consistent topic over a long span and reduces repeated content, showing better content coherence in this example. Overall, \method{} effectively mitigates repetition and topic drift in long-text generation, but there is still room to improve its overall semantic coherence.

\section{Conclusion}
We propose \method{}, a training-free length extrapolation method for long-text generation with continuous diffusion language models. \method{} combines local attention, remote context access, positional remapping, conditioning attention correction and adaptive window scheduling, and effectively mitigates the quality degradation that arises when the generation length exceeds the training length. Experiments show that, compared with conventional sliding-window attention, \method{} reduces generative perplexity and text repetition in long-text generation and improves generation quality; on the smallest backbone, the lower perplexity comes with somewhat higher repetition.

Ablation studies further confirm the effectiveness of each component. Conditioning attention correction and adaptive window scheduling mainly improve generative perplexity, while remote context access further reduces text repetition and improves the stability and coherence of long texts.

\label{end:main}
\clearpage
\section*{Limitations}
The main limitation is that we have not extensively evaluated \method{} on large-scale continuous diffusion language models. Although the experimental results are promising, its effectiveness and efficiency on larger models remain unclear. Future work will investigate the scalability and generalizability of \method{} to larger pretrained models.

\section*{Ethics Statement}
\method{} is a training-free method that modifies the attention mechanism of pretrained language models without updating their parameters. However, it may inherit existing biases, factual inaccuracies, and harmful content generation risks from the underlying models. Extending generation length may also increase the potential impact of these issues. Therefore, further safety evaluations and human oversight are necessary before applying the method in sensitive or high-stakes scenarios.

\FloatBarrier
\bibliography{refs}

\raggedbottom
\appendix
\setlength{\floatsep}{10pt}
\setlength{\textfloatsep}{12pt}
\setlength{\intextsep}{10pt}
\renewcommand{\topfraction}{.95}
\renewcommand{\bottomfraction}{.9}
\renewcommand{\textfraction}{.05}
\renewcommand{\floatpagefraction}{.65}
\setcounter{topnumber}{4}
\setcounter{bottomnumber}{3}
\setcounter{totalnumber}{5}
\renewcommand{\dbltopfraction}{.95}
\renewcommand{\dblfloatpagefraction}{.7}
\setcounter{dbltopnumber}{4}
\makeatletter
\setlength{\@fptop}{0pt}
\setlength{\@fpsep}{12pt plus 2pt}
\setlength{\@fpbot}{0pt plus 1fil}
\makeatother
\clearpage
\section{Implementation Details}
\label{app:impl}
\subsection{Models and Sampling}
\label{app:sampling}
We use the official ELF-B/M/L checkpoints without additional training. Each configuration generates 1000 samples with seed 1001 and batch size 16. ELF-B uses 32 sampling steps and ELF-M/L use 64, and all compared methods share the same initial noise.

The main hyperparameters of \method{} are a local window $W=256$, summary block size $G=16$, near range $H=512$ and $K=8$ retrieved blocks. The window is 1023 tokens early in denoising, 256 tokens later, and 2047 tokens for the final decoding call. \lite{} removes the far summaries and retrieval, keeping only the near summaries and the necessary corrections to reduce computational cost.

\subsection{Baselines}
\label{app:baselines}
We compare against five methods:
\begin{itemize}\setlength{\itemsep}{0pt}
\item \textbf{Full}: the original full attention.
\item \textbf{PI}: linearly compresses the positions of long sequences.
\item \textbf{NTK}: adjusts the RoPE frequencies to accommodate longer sequences.
\item \textbf{Self-Extend*}: keeps native positions nearby, maps distant positions in groups, and adds the conditioning corrections required by ELF.
\item \textbf{SWA$(256,4)$} \citep{jolicoeur2026sliding}: keeps the nearest 252 tokens and the first 4 body tokens.
\end{itemize}
In Table~\ref{tab:elf}, Full is the original checkpoint and the rows marked with $\rightarrow$ are its training-free variants; at 1k, the training length, the positional baselines coincide with Full.
No method updates model parameters. On CoLa and LangFlow, Table~\ref{tab:transfer} also includes SumSWA, a 256-token sliding window with mean summaries of every remaining 16-token block and without our positional or conditioning corrections.

\subsection{RangeSWA Implementation}
\label{app:rangeswaimpl}
\method{} modifies only the self-attention computation and leaves the original sampling and decoding procedure unchanged (Algorithm~\ref{alg:rangeswa}). For each body token, the local window, remote summaries, dynamically retrieved tokens and control tokens are scored together in a single softmax. A remote summary averages the keys and values of each 16-token block and receives a block-size correction. Dynamic retrieval selects up to eight relevant remote blocks of 64 tokens each; retrieved blocks use the original tokens, and overlapping summaries are removed. Retrieval is recomputed in every layer at every network call, following the changing denoising state.

\subsection{Positional and Conditioning Correction}
\label{app:poscorr}
When the generation length exceeds the training length $T=1024$, \method{} applies two corrections.

\paragraph{Positional correction.} Interactions within distance 512 keep their native positions; farther interactions are mapped in groups so that the positional distance stays within the training range.

\paragraph{Conditioning correction.} The body positions used in control--body interactions are compressed into the training range, and the attention scores from control queries to body keys are reduced by $\log(L/T)$, which mitigates the effect of the increased length on attention allocation.

When the generation length does not exceed 1024, these extrapolation corrections are disabled.

\subsection{Backbone Adaptations}
\label{app:backbones}
\paragraph{CoLa.} We modify only the attention of the latent-variable prior \citep{guo2026cola}, keeping the original variational autoencoder (VAE), causal structure and sampling procedure. Experiments use 95 continuations from 39 WikiText-103 articles, each generated after a 1k prompt and scored with conditional PPL; the reference continuations score \ColaHumanPPL{} / \ColaHumanH{} / \ColaHumanRep{} (PPL / $H$ / $R_4$\%). Because the continuations are short, CoLa's entropy is not comparable to LangFlow's document entropy.\label{app:colalong} Because the 2k prompt lies outside the positions the model was trained on, its results are not used as primary evidence of the method's effectiveness.

\paragraph{LangFlow.} We use the official OpenWebText checkpoint and the original 1024-step Euler sampler of EDM (Elucidating the Design Space of Diffusion-Based Generative Models) \citep{chen2026langflow}. The position computation is adjusted to match its RoPE implementation, and the windows for early denoising, late denoising and final readout are 1023, 256 and 2047. Each configuration generates 1000 samples, scored with unconditional Gen.PPL.

\section{Evaluation Metrics}
\label{app:metrics}
\subsection{Generation Quality Metrics}
We use three metrics to evaluate the quality of generated text.

\paragraph{Generative perplexity (Gen.PPL).} We compute the perplexity of generated text with GPT-2 Large; lower is better. Long texts are scored in blocks of at most 1024 tokens and aggregated by the number of scored tokens. ELF and LangFlow use unconditional perplexity, and CoLa uses perplexity conditioned on the prompt.

\paragraph{Unigram entropy ($H$).} $H$ measures the lexical diversity of generated text. A very low value may indicate that the model keeps repeating a small set of words. For ELF, we mark results with $H<5$ as low diversity but retain their evaluation data. In the tables, such results are shown with a red strikethrough and excluded from best-value highlighting; bold and underline mark the best and second-best of the remaining values.

\paragraph{Four-gram repetition ($R_4$).} $R_4$ measures how many four-word sequences in a text are repeated; lower means less repetition. Texts shorter than four words receive zero.

The three metrics must be read together, because a lower perplexity does not necessarily mean better generation quality and may instead come from heavily repeated text.

\subsection{Long-Range Lexical Continuity}
\label{app:similarity}
To measure how strongly the beginning and later parts of a long text are related, we use the lexical continuity metric $C_2$. From the first 500 samples with at least 240 words, we remove stop words and very short words and split each text into 60-word segments. We compare the first and third segments with term frequency--inverse document frequency (TF-IDF) cosine similarity, using the similarity between segments of different documents as a reference. A higher $C_2$ indicates a stronger lexical link between earlier and later segments.

\section{Robustness Analysis}
\label{app:robust}
\label{app:seeds}
To verify the stability of the results, we use three random seeds (1001, 1002 and 1003), generate 1000 samples per seed, and report the mean and standard deviation (Table~\ref{tab:seedsall}); --- marks configurations that were not run. \method{} consistently outperforms SWA on ELF-M/L, with small variation across seeds. On ELF-B, \method{} has lower perplexity but higher repetition than SWA.

\section{Additional Diagnostic Analysis}
\label{app:diag}
This appendix gives the protocols behind Figures~\ref{fig:extrapolation} and~\ref{fig:truncation}. The diagnostics are descriptive; component contributions are tested in Appendix~\ref{app:revision}.

\subsection{Conditioning Attention}
\label{app:diagcond}
\paragraph{Diagnostic protocol.} Figure~\ref{fig:truncation}b uses fixed denoiser states of four inputs (seed 9101) at eight denoising times per model, with the first four body tokens and all twelve control tokens retained. For each head, layer and time, control attention mass is averaged over body queries and inputs, and the figure shows the per-layer median of the SWA/Full ratio. Figure~\ref{fig:truncation}c reports LangFlow generation at 1k (1000 samples, seed 1001, 1024 steps). The same states without retained body tokens, the remote-readout cosine and the LangFlow readout error are in Figure~\ref{fig:diagdetail}. The states are correlated, so only descriptive statistics are reported. Because control keys carry no RoPE, for $L>T$ control--body scores also depend on body positions outside the training range, which the conditioning correction (Section~\ref{sec:poscond}) addresses.

\subsection{Remote Content and Length Trends}
\label{app:diagremote}
\begin{figure*}[!tp]
\centering\includegraphics[width=\textwidth]{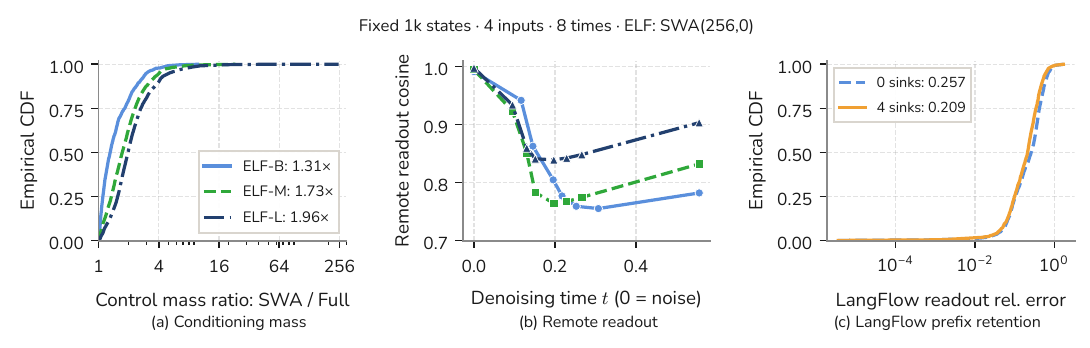}
\caption{\textbf{Fixed-state diagnostics at 1k.} (a) Control attention mass ratio and (b) summary--readout cosine on ELF-B/M/L with a 256-token window and no retained body tokens. (c) LangFlow readout error with and without the first four body tokens.}
\label{fig:diagdetail}
\end{figure*}

\begin{figure*}[!tp]
\centering\includegraphics[width=\textwidth]{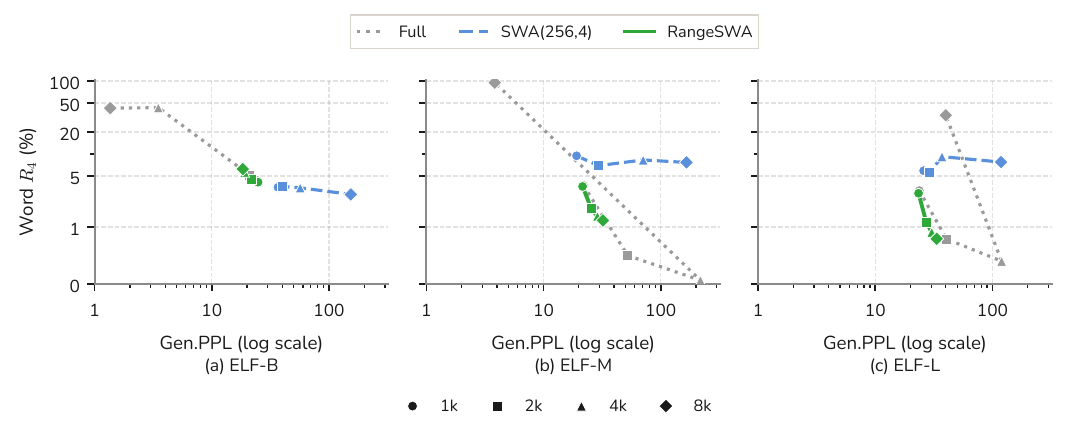}
\caption{\textbf{Perplexity--repetition trajectories over 1k--8k.} Each point is 1000 samples, and lines connect lengths. Full reaches the lowest PPL through heavy repetition at B4k/8k and M8k.}
\label{fig:trajectory}
\end{figure*}

At native length, adding block summaries to the window brings ELF-B perplexity close to Full, yet the TF-IDF similarity between distant segments and the remote-readout cosine (Figure~\ref{fig:diagdetail}) stay below Full. Figure~\ref{fig:trajectory} traces Gen.PPL and $R_4$ from 1k to 8k: Full degenerates on all three ELF models at 8k, and rescaling positions alone (PI, NTK, Self-Extend*) does not restore stable generation. Figure~\ref{fig:diagdetail} uses SWA(256,0), without retained body tokens, unlike the SWA(256,4) baseline; its four inputs and eight time points follow the protocol above, and the LangFlow readout error in panel (c) does not prevent degenerate generation (Figure~\ref{fig:truncation}c). In Figures~\ref{fig:extrapolation}b and~\ref{fig:trajectory}, PPL uses a logarithmic axis; in Figure~\ref{fig:trajectory}, $R_4$ is linear below $1\%$ and logarithmic above. The metrics in Figure~\ref{fig:extrapolation}a are aggregated over 1000 samples, and Figure~\ref{fig:case} scores PPL with GPT-2 Large in 1024-token chunks.

\section{Component Ablation Studies}
\label{app:revision}
To analyze the role of each component of \method{}, we remove the positional correction, the conditioning correction, the remote summaries, dynamic retrieval and the adaptive window in turn, and compare generative perplexity (PPL), repetition ($R_4$) and lexical continuity ($C_2$). The experiments are mainly run on ELF-B at 2k, 4k and 8k and verified on ELF-M/L at 8k. Each configuration generates 1000 samples with the same random seed.\label{app:sensitivity} All parameters were fixed on ELF-B first and then used directly for the other models without further tuning, so the ELF-B results are not independent of this choice. Table~\ref{tab:ablationmain} in the main text deletes one component at a time from \method{} on ELF-B at 8k (1000 outputs, seed 1001); there the mass bias acts only on control queries, native far positions retain the density bias, and native full decoding replaces the complete decoder readout.

\input{generated/table_components.tex}

\subsection{Positional and Conditioning Corrections}
Tables~\ref{tab:components}, \ref{tab:mechanismrevision} and~\ref{tab:mechanismtransfer} show the effect of the positional and conditioning corrections. Removing the conditioning position interpolation clearly increases perplexity: on ELF-B at 8k, PPL rises from 18.37 to 76.68, and on ELF-M/L at 8k from 32.13 and 33.39 to 164.90 and 61.42. Restoring the native far-body positions lowers PPL on ELF-B, but raises the 8k repetition rate from 6.24\% to 9.45\%, so the positional correction trades perplexity against repetition. In contrast, removing the attention-mass correction of control queries has a smaller effect, although it still clearly increases perplexity on ELF-M.

\subsection{Remote Summaries and Dynamic Retrieval}
Tables~\ref{tab:components}--\ref{tab:mechanismtransfer} further analyze the role of remote information. On ELF-B, removing either the summaries or dynamic retrieval increases perplexity; at 8k, PPL rises from 18.37 to 22.15 without summaries and to 21.51 without retrieval. Table~\ref{tab:ablation} also shows that removing dynamic retrieval at 2k lowers lexical continuity $C_2$ from 0.144 to 0.065, indicating that retrieval helps preserve long-range lexical links. In the ELF-M/L 8k experiments, removing retrieval lowers PPL but raises the repetition rate from 1.23\% and 0.79\% to 3.70\% and 1.83\%. These results show that remote summaries and dynamic retrieval play different roles, and dynamic retrieval in particular helps reduce text repetition on ELF-M/L.

\input{generated/table_mechanism_revision.tex}

\subsection{Adaptive Window Schedule}
Tables~\ref{tab:components} and~\ref{tab:ablation} show the role of the window schedule. Removing the large early-denoising window raises the ELF-B 8k PPL from 18.37 to 22.08 and lowers $C_2$ from 0.084 to 0.036, indicating that a wider context early in denoising improves generation. Restoring native full attention in the final decoding call raises the ELF-B 8k PPL to 80.43, while the 2k result changes little. Both the denoising-stage window adjustment and the bounded final decoding window are therefore important for long-text generation.

\begin{table*}[!tp]
\centering\small\input{generated/table_ablation.tex}
\caption{\textbf{Ablations of remote information and the window schedule on ELF-B.} Lexical continuity $C_2$ is shown with its cross-document reference.}
\label{tab:ablation}
\end{table*}

\input{generated/table_mechanism_transfer.tex}

\subsection{Parameter Sensitivity}
Table~\ref{tab:hypersensitivity} tests the local window size $W$, summary block size $G$, number of retrieved blocks $K$ and the early-window switch threshold. The experiments are run on ELF-B at 4k, changing one parameter at a time. Different parameters affect perplexity and repetition; for example, increasing $K$ from 8 to 16 lowers PPL from 19.06 to 17.31 but raises the repetition rate from 5.64\% to 7.23\%. Balancing generation quality and repetition, we keep the default setting fixed before this sweep, $W=256$, $G=16$, $K=8$ and a window-switch threshold of 0.182, and use it for the other models.

\input{generated/table_hyperparameter_sensitivity.tex}

\section{Component-wise Validation with Simplified Variants}
\label{app:enhancedswa}
To further analyze the role of each component of \method{}, we add the conditioning corrections and the adaptive window schedule step by step to a plain sliding window and compare the result with the complete \method{}. The experiments use the same models and sampling settings as the main text, with 1000 samples per configuration. None of the simplified variants uses remote summaries, dynamic retrieval or grouped far positions. In Tables~\ref{tab:remote}, \ref{tab:enhancedswakeys} and~\ref{tab:enhancedswa}, $^\dagger$ marks these simplified variants of \method{}, C denotes the conditioning corrections and S the window schedule. B1 also bounds the decoder window, and SWA$_{1280}$ (B0) matches the key count of \method{}, not its wall-clock cost.

\subsection{Conditioning Corrections and Window Schedule}
\label{app:ladderA}
We design four simplified variants to compare the effect of each component on generation quality:
\begin{itemize}\setlength{\itemsep}{0pt}
\item \textbf{A0}: a plain 256-token sliding window.
\item \textbf{A1}: A0 plus the conditioning attention corrections.
\item \textbf{A2}: A0 plus the adaptive window schedule.
\item \textbf{A3}: A0 plus both the conditioning corrections and the window schedule.
\end{itemize}
A0 differs slightly from SWA(256,4) in the main text: it does not retain the first 4 body tokens, so it serves as the common reference for this set of experiments. As shown in Table~\ref{tab:enhancedswa}, at 8k A3 lowers the PPL of ELF-B/M/L from 171.70, 158.71 and 121.22 to 28.95, 24.31 and 28.89. In contrast, A2, which adds only the window schedule, does not lower PPL consistently. Combining the conditioning corrections with the window schedule is therefore more effective.

\subsection{Comparison at a Matched Attention Budget}
\label{app:ladderB}
To rule out the effect of attention window size, we further compare methods that use a similar number of keys. At 8k, each \method{} body query scores about
\[
256+8\times64+\frac{2\times512}{16}+\frac{8192-1024}{16}=1280
\]
keys: a local window of 256, 8 retrieved blocks of 64 tokens, near summaries of the two 512-token flanks, and far summaries of the remaining $8192-1024$ positions, each summary covering 16 tokens. We therefore set:
\begin{itemize}\setlength{\itemsep}{0pt}
\item \textbf{B0}: plain SWA with a 1280-token window.
\item \textbf{B1}: B0 plus the conditioning corrections and the bounded decoding window.
\end{itemize}
As shown in Table~\ref{tab:enhancedswakeys}, in the ELF-B/M/L 8k experiments B0 has PPL 162.11, 133.51 and 92.97, whereas B1 lowers it to 24.01, 27.34 and 29.85. Simply enlarging the attention window therefore does not solve long-text degeneration; the conditioning corrections and the bounded decoding window are also important.

\input{generated/table_enhanced_swa_keys.tex}

\subsection{Results}
\label{app:enhancedread}
Tables~\ref{tab:enhancedswakeys} and~\ref{tab:enhancedswa} show that the components of \method{} are complementary. First, the conditioning corrections and window schedule provide most of the perplexity improvement: in the ELF-M/L 8k experiments, the simplified variant A3 even reaches lower PPL than the complete \method{}, 24.31 and 28.89 versus 32.13 and 33.39. Second, the complete \method{} has an advantage in reducing repetition: at ELF-M/L 8k its $R_4$ is 1.23\% and 0.79\%, lower than A3's 3.60\% and 1.94\%. On ELF-B, however, the complete \method{} has lower PPL but higher repetition than these simplified variants. Overall, the conditioning corrections and window schedule mainly improve perplexity, while remote summaries and dynamic retrieval further reduce repetition on ELF-M/L, and the effect of each component differs across models. All rungs use one seed.

\input{generated/table_enhanced_swa.tex}

\section{Inference Efficiency}
\label{app:eff}
To evaluate the computational efficiency of \method{}, we implement its attention in Triton and measure running time and generation quality on ELF-B at different generation lengths. The experiments use a single H100 GPU with batch size 16, 32 sampling steps and the final decoding call, in both BF16 and strict 32-bit floating-point (FP32) precision.

\subsection{Triton Implementation}
\label{app:triton}
The Triton kernel merges the attention over the local window, remote summaries, dynamically retrieved tokens and control tokens into a single softmax. Each network call computes the position-encoded keys and the remote summaries only once, avoiding repeated computation, and each position range uses its corresponding position mapping. Retrieval selection and the control-token queries still use the reference implementation. Table~\ref{tab:kernelprofile} splits the running time of one denoising call into attention, preparation and the remaining network computation. In BF16 at 8k, the attention time of \method{} is 72.7\,ms, lower than the 126.4\,ms of Full, showing that sparse attention reduces computational cost on long sequences.

\begin{table}[!tp]
\centering\small\input{generated/table_kernel_v2_profile.tex}
\caption{\textbf{Time breakdown of one ELF-B denoising call.} Attention, preparation and remaining network time in milliseconds, in BF16 and FP32.}
\label{tab:kernelprofile}
\end{table}

\subsection{Inference Speed}
\label{app:speed}
Table~\ref{tab:efficiency} (main text, BF16) and Table~\ref{tab:efficiencyfull} (BF16 and strict FP32) report the complete sampling time at different generation lengths. Each configuration is run five times; we report the median time and the median paired speedup over Full, with shared initial noise and rotated method order. BF16 Full uses FlashAttention and strict FP32 Full uses PyTorch scaled dot-product attention (SDPA). In BF16, \method{} is slightly slower than Full at 2k, about equally fast at 4k, and 1.21$\times$ faster at 8k, where sampling time drops from 10.84\,s to 8.98\,s. \lite{} further reduces computation and reaches 1.15$\times$, 1.32$\times$ and 1.62$\times$ speedups at 2k, 4k and 8k. In strict FP32, \method{} and \lite{} reach 1.15$\times$ and 2.18$\times$ speedups at 8k. The efficiency advantage of \method{} therefore appears mainly for longer generations, and the Lite version has a lower computational cost.

\begin{table*}[t]
\centering\setlength{\tabcolsep}{2.5pt}\resizebox{\textwidth}{!}{\input{generated/table_kernel_v2.tex}}
\caption{\textbf{Sampling time and speedup over Full.} ELF-B on one H100 in BF16 and strict FP32; median of five runs.}
\label{tab:efficiencyfull}
\end{table*}

\input{generated/table_triton2_quality.tex}

\subsection{Numerical Consistency and Generation Quality}
\label{app:numerics}
To confirm that the Triton implementation does not noticeably change model outputs, we run numerical error checks and generation-quality tests. As shown in Table~\ref{tab:tritonquality2}, the maximum per-layer relative errors of \method{} and \lite{} in the strict FP32 check are $1.71\times10^{-6}$ and $1.07\times10^{-6}$, both below $10^{-5}$. Each setting generates 1000 samples, and we compare PPL, $H$ and $R_4$. The generation quality of the Triton implementation is essentially the same as that of the corresponding reference implementation, with very small PPL differences. The Triton implementation therefore improves long-text generation efficiency while preserving numerical consistency and generation quality.

\section{Model-Based Evaluation}
\label{app:evaluation}
To further evaluate the semantic coherence of long texts, we score generated texts automatically with Qwen3-14B. All evaluations are performed automatically by a language model run by the authors; no human rating is involved, and we report no independent human evaluation.

\label{app:judge14}We use Qwen3-14B as the judge with greedy decoding and the thinking mode off. Each configuration compares the first 200 generated samples with the same IDs. The judge focuses on whether the topic is consistent, whether entities and events are coherent, and whether there is repeated or irrelevant content, and it is instructed not to give higher scores to longer texts. To reduce order bias, we swap the presentation order of the two texts and evaluate twice; a win or loss is recorded only when both judgments agree, and otherwise the pair counts as a tie. For texts longer than 6000 words, only the first and last 3000 words are kept, and each end is further limited to 5000 judge tokens.

Table~\ref{tab:llmjudge} reports the win, tie and loss rates (W/T/L) of \method{} in percent of 200 pairs, the net win rate W$-$L with its 95\% confidence interval, and single-document coherence scores on a 1--5 scale, with $\Delta$ the score of \method{} minus that of Full. When \method{} wins every pair the interval degenerates at $+100$ and is shown as ---. At 4k and 8k, \method{} clearly outperforms SWA and Full in the pairwise comparisons; at 8k in particular, \method{} wins every pairwise evaluation on all three ELF models. The single-document scores remain low, however: \method{} scores about 2.0--2.5 at 8k. At 2k, Full receives slightly higher single-document scores than \method{}, and the difference is clear only on ELF-B. These results indicate that \method{} mitigates long-text degeneration, but its overall semantic coherence still leaves room for improvement.

\input{generated/table_llm_judge.tex}

\end{document}

%% file: generated/kernel_v2_macros.tex
\newcommand{\SpRangeBFEight}{1.21$\times$}
\newcommand{\SpLiteBFEight}{1.62$\times$}

%% file: generated/cola_macros.tex
\newcommand{\ColaHumanPPL}{15.95}

\newcommand{\ColaHumanH}{4.28}
\newcommand{\ColaHumanRep}{0.13}

%% file: generated/table1_elf.tex
{\lcstyle\begin{tabular}{l|ccc|ccc|ccc|ccc}
\toprule[1pt]
 & \multicolumn{3}{c|}{1k} & \multicolumn{3}{c|}{2k} & \multicolumn{3}{c|}{4k} & \multicolumn{3}{c}{8k} \\
\cmidrule(lr){2-4}\cmidrule(lr){5-7}\cmidrule(lr){8-10}\cmidrule(lr){11-13}
Method & PPL$\downarrow$ & $H$ & $R_4$$\downarrow$ & PPL$\downarrow$ & $H$ & $R_4$$\downarrow$ & PPL$\downarrow$ & $H$ & $R_4$$\downarrow$ & PPL$\downarrow$ & $H$ & $R_4$$\downarrow$ \\
\midrule
ELF-B Full & \textbf{24.42} & 5.15 & 4.4 & \textcolor{red}{\sout{21.10}} & \textcolor{red}{\sout{4.97}} & \textcolor{red}{\sout{5.1}} & \textcolor{red}{\sout{3.49}} & \textcolor{red}{\sout{1.91}} & \textcolor{red}{\sout{43.1}} & \textcolor{red}{\sout{1.36}} & \textcolor{red}{\sout{0.94}} & \textcolor{red}{\sout{42.6}} \\
$\rightarrow$ Full + PI & \textbf{24.42} & 5.15 & 4.4 & 78.28 & 5.13 & 13.2 & \textcolor{red}{\sout{10.77}} & \textcolor{red}{\sout{3.83}} & \textcolor{red}{\sout{49.8}} & \textcolor{red}{\sout{1.62}} & \textcolor{red}{\sout{2.05}} & \textcolor{red}{\sout{76.1}} \\
$\rightarrow$ Full + NTK & \textbf{24.42} & 5.15 & 4.4 & \textcolor{red}{\sout{16.68}} & \textcolor{red}{\sout{4.23}} & \textcolor{red}{\sout{12.5}} & \textcolor{red}{\sout{8.67}} & \textcolor{red}{\sout{2.36}} & \textcolor{red}{\sout{57.9}} & \textcolor{red}{\sout{1.72}} & \textcolor{red}{\sout{0.55}} & \textcolor{red}{\sout{15.3}} \\
$\rightarrow$ Self-Extend* & \textbf{24.42} & 5.15 & 4.4 & \textbf{20.05} & 5.01 & 7.0 & \textcolor{red}{\sout{7.97}} & \textcolor{red}{\sout{2.75}} & \textcolor{red}{\sout{48.8}} & \textcolor{red}{\sout{5.90}} & \textcolor{red}{\sout{1.89}} & \textcolor{red}{\sout{82.7}} \\
$\rightarrow$ SWA(256,4) & 36.68 & 5.35 & \textbf{3.5} & 40.24 & 5.75 & \textbf{3.6} & \underline{56.79} & 6.06 & \textbf{3.4} & \underline{153.55} & 5.56 & \textbf{2.8} \\
$\rightarrow$ RangeSWA (Ours) & \underline{24.80} & 5.18 & \underline{4.1} & \underline{21.58} & 5.36 & \underline{4.6} & \textbf{19.06} & 5.51 & \underline{5.6} & \textbf{18.37} & 5.63 & \underline{6.2} \\
\midrule
ELF-M Full & 21.91 & 5.19 & \underline{3.7} & 52.40 & 5.52 & \textbf{0.5} & 218.21 & 5.34 & \textbf{0.1} & \textcolor{red}{\sout{3.84}} & \textcolor{red}{\sout{0.80}} & \textcolor{red}{\sout{95.1}} \\
$\rightarrow$ Full + PI & 21.91 & 5.19 & \underline{3.7} & 272.32 & 5.67 & 1.6 & \textcolor{red}{\sout{207.72}} & \textcolor{red}{\sout{4.67}} & \textcolor{red}{\sout{0.3}} & \textcolor{red}{\sout{58.05}} & \textcolor{red}{\sout{4.55}} & \textcolor{red}{\sout{24.0}} \\
$\rightarrow$ Full + NTK & 21.91 & 5.19 & \underline{3.7} & 45.17 & 5.25 & \underline{1.1} & \textcolor{red}{\sout{111.71}} & \textcolor{red}{\sout{3.96}} & \textcolor{red}{\sout{0.7}} & \textcolor{red}{\sout{48.51}} & \textcolor{red}{\sout{3.19}} & \textcolor{red}{\sout{23.5}} \\
$\rightarrow$ Self-Extend* & 21.91 & 5.19 & \underline{3.7} & 31.06 & 5.37 & 1.5 & 72.56 & 5.21 & \underline{0.4} & \textcolor{red}{\sout{94.06}} & \textcolor{red}{\sout{4.01}} & \textcolor{red}{\sout{1.1}} \\
$\rightarrow$ SWA(256,4) & \textbf{19.13} & 5.21 & 9.5 & \underline{29.40} & 5.78 & 6.9 & \underline{70.81} & 6.27 & 8.3 & \underline{165.93} & 6.50 & \underline{7.7} \\
$\rightarrow$ RangeSWA (Ours) & \underline{21.52} & 5.20 & \textbf{3.6} & \textbf{25.55} & 5.56 & 1.8 & \textbf{28.82} & 5.91 & 1.4 & \textbf{32.13} & 6.21 & \textbf{1.2} \\
\midrule
ELF-L Full & \underline{23.74} & 5.28 & \underline{3.2} & 40.56 & 5.58 & \textbf{0.8} & \textcolor{red}{\sout{119.68}} & \textcolor{red}{\sout{4.50}} & \textcolor{red}{\sout{0.4}} & \textcolor{red}{\sout{39.93}} & \textcolor{red}{\sout{3.49}} & \textcolor{red}{\sout{34.0}} \\
$\rightarrow$ Full + PI & \underline{23.74} & 5.28 & \underline{3.2} & 211.33 & 5.71 & 4.3 & 202.34 & 5.26 & \underline{3.7} & \textcolor{red}{\sout{92.76}} & \textcolor{red}{\sout{3.95}} & \textcolor{red}{\sout{2.8}} \\
$\rightarrow$ Full + NTK & \underline{23.74} & 5.28 & \underline{3.2} & 44.01 & 5.20 & 1.9 & \textcolor{red}{\sout{45.62}} & \textcolor{red}{\sout{3.51}} & \textcolor{red}{\sout{5.1}} & \textcolor{red}{\sout{39.24}} & \textcolor{red}{\sout{3.12}} & \textcolor{red}{\sout{18.1}} \\
$\rightarrow$ Self-Extend* & \underline{23.74} & 5.28 & \underline{3.2} & 32.80 & 5.56 & 1.3 & \textcolor{red}{\sout{79.74}} & \textcolor{red}{\sout{4.85}} & \textcolor{red}{\sout{1.6}} & \textcolor{red}{\sout{56.56}} & \textcolor{red}{\sout{3.52}} & \textcolor{red}{\sout{5.6}} \\
$\rightarrow$ SWA(256,4) & 25.85 & 5.21 & 5.9 & \underline{29.09} & 5.69 & 5.5 & \underline{37.05} & 6.08 & 9.2 & \underline{118.31} & 6.46 & \underline{7.8} \\
$\rightarrow$ RangeSWA (Ours) & \textbf{23.39} & 5.29 & \textbf{2.9} & \textbf{27.27} & 5.68 & \underline{1.2} & \textbf{30.30} & 6.05 & \textbf{0.9} & \textbf{33.39} & 6.36 & \textbf{0.8} \\
\bottomrule[1pt]
\end{tabular}}

%% file: generated/table_seeds_all.tex
\begin{table}[!htbp]
\centering\small\setlength{\tabcolsep}{3pt}
\begin{adjustbox}{max width=\linewidth}
{\lcstyle\begin{tabular}{l|cccc}
\toprule[1pt]
 & \multicolumn{2}{c}{SWA(256,4)} & \multicolumn{2}{c}{\method{}} \\
\cmidrule(lr){2-3}\cmidrule(lr){4-5}
Length & PPL & $R_4$(\%) & PPL & $R_4$(\%) \\
\midrule
\textbf{ELF-B} &  &  &  &  \\
1024 & \appending{---} & \appending{---} & $24.57\pm0.21$ & $4.13\pm0.02$ \\
2048 & \apsecond{$41.11\pm0.80$} & \apbest{$3.43\pm0.14$} & \apbest{$21.70\pm0.12$} & \apsecond{$4.50\pm0.05$} \\
4096 & \apsecond{$57.15\pm0.64$} & \apbest{$3.35\pm0.09$} & \apbest{$19.08\pm0.17$} & \apsecond{$5.49\pm0.21$} \\
8192 & \apsecond{$153.12\pm1.45$} & \apbest{$2.85\pm0.04$} & \apbest{$18.27\pm0.09$} & \apsecond{$6.29\pm0.08$} \\
\midrule
\textbf{ELF-M} &  &  &  &  \\
2048 & \apsecond{$29.60\pm0.17$} & \apsecond{$6.81\pm0.12$} & \apbest{$25.53\pm0.11$} & \apbest{$1.78\pm0.01$} \\
4096 & \apsecond{$71.35\pm0.79$} & \apsecond{$8.24\pm0.17$} & \apbest{$28.96\pm0.31$} & \apbest{$1.38\pm0.03$} \\
8192 & \apsecond{$166.76\pm0.73$} & \apsecond{$7.70\pm0.08$} & \apbest{$32.27\pm0.15$} & \apbest{$1.21\pm0.02$} \\
\midrule
\textbf{ELF-L} &  &  &  &  \\
2048 & \apsecond{$28.96\pm0.12$} & \apsecond{$5.64\pm0.11$} & \apbest{$27.09\pm0.19$} & \apbest{$1.18\pm0.02$} \\
4096 & \apsecond{$37.59\pm0.69$} & \apsecond{$9.04\pm0.37$} & \apbest{$30.43\pm0.38$} & \apbest{$0.90\pm0.03$} \\
8192 & \apsecond{$119.12\pm0.78$} & \apsecond{$7.72\pm0.05$} & \apbest{$33.74\pm0.32$} & \apbest{$0.77\pm0.02$} \\
\bottomrule[1pt]
\end{tabular}}
\end{adjustbox}
\apnote{Within model and length: mean PPL $\downarrow$, mean $R_4\downarrow$; SD retained. Rows with one method unranked; --- denotes configurations not run.}

\caption{\textbf{Three-seed robustness on ELF-B/M/L.} Mean $\pm$ standard deviation (SD) over seeds 1001--1003, 1000 outputs per seed.}
\label{tab:seedsall}
\end{table}

%% file: generated/table_transfer_main.tex
{\lcstyle\begin{tabular}{l|ccc|ccc|ccc}
\toprule[1pt]
 & \multicolumn{3}{c|}{CoLa: 1k prompt} & \multicolumn{3}{c|}{LangFlow: 1k} & \multicolumn{3}{c}{LangFlow: 2k} \\
\cmidrule(lr){2-4}\cmidrule(lr){5-7}\cmidrule(lr){8-10}
Method & PPL$\downarrow$ & $H$ & $R_4$$\downarrow$ & PPL$\downarrow$ & $H$ & $R_4$$\downarrow$ & PPL$\downarrow$ & $H$ & $R_4$$\downarrow$ \\
\midrule
Full & 80.92 & 3.57 & 1.3 & 36.17 & 5.25 & 4.9 & 99.36 & 5.86 & 0.2 \\
SWA & 20.32 & 3.61 & 7.3 & 5.61 & 4.25 & 51.1 & 6.09 & 4.86 & 49.5 \\
SumSWA & 28.61 & 3.61 & 4.7 & 15.63 & 5.08 & 23.3 & 42.75 & 5.75 & 5.3 \\
RangeSWA (Ours) & 56.50 & 3.59 & 1.6 & 36.97 & 5.27 & 4.3 & 33.94 & 5.68 & 5.3 \\
\bottomrule[1pt]
\end{tabular}}

%% file: generated/table_ablation_main.tex
{\lcstyle\begin{tabular}{l|rr}
\toprule[1pt]
Intervention & Gen.PPL & $R_4$(\%) \\
\midrule
RangeSWA (Ours) & 18.37 & 6.24 \\
No conditioning PI & 76.68 & 2.55 \\
No control mass bias & 19.33 & 5.61 \\
Native far positions & 13.67 & 9.45 \\
No early wide window & 22.08 & 4.73 \\
Native full decoder & 80.43 & 3.85 \\
\bottomrule[1pt]
\end{tabular}}

%% file: generated/table_ablation_remote.tex
{\lcstyle\begin{tabular}{l|cc|cc|cc}
\toprule[1pt]
 & \multicolumn{2}{c|}{ELF-B} & \multicolumn{2}{c|}{ELF-M} & \multicolumn{2}{c}{ELF-L} \\
\cmidrule(lr){2-3}\cmidrule(lr){4-5}\cmidrule(lr){6-7}
Variant & PPL$\downarrow$ & $R_4$$\downarrow$ & PPL$\downarrow$ & $R_4$$\downarrow$ & PPL$\downarrow$ & $R_4$$\downarrow$ \\
\midrule
RangeSWA (Ours) & 18.37 & 6.24 & 32.13 & 1.23 & 33.39 & 0.79 \\
w/o retrieval & 21.51 & 6.21 & 23.06 & 3.70 & 25.92 & 1.83 \\
A3: SWA+C+S$^\dagger$ & 28.95 & 4.64 & 24.31 & 3.60 & 28.89 & 1.94 \\
B0: SWA$_{1280}$ & 162.11 & 4.16 & 133.51 & 1.87 & 92.97 & 3.07 \\
B1: SWA$_{1280}$+C$^\dagger$ & 24.01 & 4.46 & 27.34 & 2.08 & 29.85 & 1.66 \\
\bottomrule[1pt]
\end{tabular}}

%% file: generated/table_efficiency_main.tex
{\lcstyle\begin{tabular}{l|cc|cc|cc}
\toprule[1pt]
 & \multicolumn{2}{c|}{2k} & \multicolumn{2}{c|}{4k} & \multicolumn{2}{c}{8k} \\
\cmidrule(lr){2-3}\cmidrule(lr){4-5}\cmidrule(lr){6-7}
Method & s$\downarrow$ & Speedup$\uparrow$ & s$\downarrow$ & Speedup$\uparrow$ & s$\downarrow$ & Speedup$\uparrow$ \\
\midrule
Full & 2.08 & -- & 4.48 & -- & 10.84 & -- \\
\method{} (Ours) & 2.36 & 0.88$\times$ & 4.46 & 1.01$\times$ & 8.98 & 1.21$\times$ \\
\lite{} (Ours) & 1.81 & 1.15$\times$ & 3.41 & 1.32$\times$ & 6.71 & 1.62$\times$ \\
\bottomrule[1pt]
\end{tabular}}

%% file: generated/case_main.tex
\begin{adjustbox}{max width=\linewidth}
\begin{tabular}{@{}p{.215\columnwidth}@{\hspace{5pt}}p{.75\columnwidth}@{}}
\toprule
Full\newline PPL 19.22\newline $H$ 4.91\newline $R_4$ 4.95\%\newline 1679 words & \textit{Start:} Start designing your blog. The goal is to make your blog audience positive. Facebook posts in \dots\newline\textit{End:} \dots{} how to make an positive impact. Use your blog or social media post as an example\newline{}[\textit{Repeated generic advice}] \\
\midrule
SWA\newline PPL 52.05\newline $H$ 5.90\newline $R_4$ 2.18\%\newline 1375 words & \textit{Start:} Arkuyi Opsuma met with Jakaki this week and expressed his satisfaction that some players have gained \dots\newline\textit{End:} \dots{} institutions from revealing important information. Summs were routered from Norway and shipped into Europe, according to\newline{}[\textit{Sports to central banking}] \\
\midrule
\textbf{RangeSWA}\newline PPL 27.48\newline $H$ 5.47\newline $R_4$ 0.73\%\newline 1498 words & \textit{Start:} One of the problems with Webtop is the fact that Webtop lacks its full user experience \dots\newline\textit{End:} \dots{} designers and developers to teach users how to make their web experience real, mobile, and comfortable..\newline{}[\textit{Webtop topic retained}] \\
\bottomrule
\end{tabular}
\end{adjustbox}
\apnote{Diagnostic values; no best/second ranking.}

%% file: generated/table_components.tex
\begin{table}[!htbp]
\centering\small\setlength{\tabcolsep}{3pt}
\begin{adjustbox}{max width=\linewidth}
{\lcstyle\begin{tabular}{l|ccc|ccc}
\toprule[1pt]
 & \multicolumn{3}{c|}{B 2k} & \multicolumn{3}{c}{B 8k} \\
\cmidrule(lr){2-4}\cmidrule(lr){5-7}
Intervention & PPL$\downarrow$ & $H$ & $R_4$$\downarrow$ & PPL$\downarrow$ & $H$ & $R_4$$\downarrow$ \\
\midrule
RangeSWA (Ours) & 21.58 & 5.36 & 4.55 & \apsecond{18.37} & 5.63 & 6.24 \\
No conditioning interpolation & 24.45 & 5.44 & \apbest{3.74} & 76.68 & 5.95 & \apbest{2.55} \\
No control-row mass correction & 21.99 & 5.37 & 4.41 & 19.33 & 5.68 & 5.61 \\
Native body-body positions & \apbest{19.63} & 5.22 & 4.87 & \apbest{13.67} & 5.34 & 9.45 \\
No summary count bias & 23.44 & 5.41 & 3.92 & 21.81 & 5.80 & 4.37 \\
No summaries & 23.59 & 5.41 & \apsecond{3.87} & 22.15 & 5.81 & 4.27 \\
No exact retrieval & 23.97 & 5.55 & 5.07 & 21.51 & 6.01 & 6.21 \\
\midrule
No early wide window & 24.26 & 5.56 & 4.02 & 22.08 & 5.99 & 4.73 \\
No bounded decoder & \apsecond{21.49} & 5.36 & 4.55 & 80.43 & 5.71 & \apsecond{3.85} \\
\bottomrule[1pt]
\end{tabular}}
\end{adjustbox}
\apnote{Within model and length: PPL $\downarrow$, $R_4\downarrow$.}

\caption{\textbf{Component ablations of ELF-B at 2k and 8k.} Each row changes one component; 1000 samples per configuration.}
\label{tab:components}
\end{table}

%% file: generated/table_mechanism_revision.tex
\begin{table*}[!tp]
\centering\scriptsize
\begin{adjustbox}{max width=\linewidth}
{\lcstyle\begin{tabular}{l|ccc|ccc|ccc}
\toprule[1pt]
 & \multicolumn{3}{c|}{2k} & \multicolumn{3}{c|}{4k} & \multicolumn{3}{c}{8k} \\
\cmidrule(lr){2-4}\cmidrule(lr){5-7}\cmidrule(lr){8-10}
ELF-B intervention & PPL$\downarrow$ & $H$ & $R_4$$\downarrow$ & PPL$\downarrow$ & $H$ & $R_4$$\downarrow$ & PPL$\downarrow$ & $H$ & $R_4$$\downarrow$ \\
\midrule
RangeSWA (Ours) & \apsecond{21.58} & 5.36 & 4.55 & \apsecond{19.06} & 5.51 & 5.64 & \apsecond{18.37} & 5.63 & 6.24 \\
No conditioning interpolation & 24.45 & 5.44 & \apbest{3.74} & 37.84 & 5.81 & \apbest{2.87} & 76.68 & 5.95 & \apbest{2.55} \\
No control-row mass correction & 21.99 & 5.37 & 4.41 & 19.71 & 5.54 & 5.32 & 19.33 & 5.68 & 5.61 \\
Native body-body positions & \apbest{19.63} & 5.22 & 4.87 & \apbest{15.69} & 5.30 & 6.85 & \apbest{13.67} & 5.34 & 9.45 \\
No summary count bias & 23.44 & 5.41 & 3.92 & 22.00 & 5.63 & 4.34 & 21.81 & 5.80 & 4.37 \\
No summaries & 23.59 & 5.41 & \apsecond{3.87} & 22.24 & 5.64 & \apsecond{4.24} & 22.15 & 5.81 & \apsecond{4.27} \\
No exact retrieval & 23.97 & 5.55 & 5.07 & 22.23 & 5.80 & 5.85 & 21.51 & 6.01 & 6.21 \\
\bottomrule[1pt]
\end{tabular}}
\end{adjustbox}
\apnote{Within length: PPL $\downarrow$, $R_4\downarrow$.}

\caption{\textbf{Single-component ablations of ELF-B at 2k--8k.} The positional and conditioning corrections, summaries and retrieval are removed one at a time.}
\label{tab:mechanismrevision}
\end{table*}

%% file: generated/table_ablation.tex
\begin{adjustbox}{max width=\linewidth}
{\lcstyle\begin{tabular}{l|ccc|cc|ccc|cc}
\toprule[1pt]
 & \multicolumn{3}{c|}{2k: PPL / $H$ / $R_4$} & \multicolumn{2}{c|}{$C_2$ / cross} & \multicolumn{3}{c|}{8k: PPL / $H$ / $R_4$} & \multicolumn{2}{c}{$C_2$ / cross} \\
\cmidrule(lr){2-4}\cmidrule(lr){5-6}\cmidrule(lr){7-9}\cmidrule(lr){10-11}
ELF-B deletion & PPL$\downarrow$ & $H$ & $R_4$$\downarrow$ & $C_2$$\uparrow$ & cross & PPL$\downarrow$ & $H$ & $R_4$$\downarrow$ & $C_2$$\uparrow$ & cross \\
\midrule
RangeSWA (Ours) & \apsecond{21.58} & 5.36 & \apsecond{4.6} & \apbest{0.144} & 0.018 & \apbest{18.37} & 5.63 & 6.2 & \apsecond{0.084} & 0.030 \\
No exact retrieval & 23.97 & 5.55 & 5.1 & \apsecond{0.065} & 0.012 & \apsecond{21.51} & 6.01 & 6.2 & 0.062 & 0.013 \\
No early wide window & 24.26 & 5.56 & \apbest{4.0} & 0.061 & 0.015 & 22.08 & 5.99 & \apsecond{4.7} & 0.036 & 0.018 \\
No bounded decoder & \apbest{21.49} & 5.36 & \apsecond{4.6} & \apbest{0.144} & 0.018 & 80.43 & 5.71 & \apbest{3.9} & \apbest{0.086} & 0.030 \\
\bottomrule[1pt]
\end{tabular}}
\end{adjustbox}
\apnote{Within length: PPL $\downarrow$, $R_4\downarrow$, $C_2\uparrow$.}

%% file: generated/table_mechanism_transfer.tex
\begin{table}[!htbp]
\centering\small
\begin{adjustbox}{max width=\linewidth}
{\lcstyle\begin{tabular}{l|ccc}
\toprule[1pt]
 & \multicolumn{3}{c}{PPL / $H$ / $R_4$(\%)} \\
\cmidrule(lr){2-4}
Variant & PPL$\downarrow$ & $H$ & $R_4$$\downarrow$ \\
\midrule
\textbf{ELF-M} &  &  &  \\
No conditioning interpolation & 164.90 & 6.49 & \apbest{0.64} \\
No control-row mass correction & \apsecond{66.92} & 6.48 & \apsecond{0.88} \\
No exact retrieval & \apbest{23.06} & 6.18 & 3.70 \\
\textbf{ELF-L} &  &  &  \\
No conditioning interpolation & 61.42 & 6.51 & \apsecond{1.71} \\
No control-row mass correction & \apsecond{33.90} & 6.37 & \apbest{1.20} \\
No exact retrieval & \apbest{25.92} & 6.34 & 1.83 \\
\bottomrule[1pt]
\end{tabular}}
\end{adjustbox}
\apnote{Within model at fixed length: PPL $\downarrow$, $R_4\downarrow$; pending and unmatched rows unranked.}

\caption{\textbf{Component ablations of ELF-M/L at 8k.} The conditioning corrections and retrieval are removed one at a time.}
\label{tab:mechanismtransfer}
\end{table}

%% file: generated/table_hyperparameter_sensitivity.tex
\begin{table}[!htbp]
\centering\small
\begin{adjustbox}{max width=\linewidth}
{\lcstyle\begin{tabular}{l|ccc}
\toprule[1pt]
 & \multicolumn{3}{c}{PPL / $H$ / $R_4$(\%)} \\
\cmidrule(lr){2-4}
Variant & PPL$\downarrow$ & $H$ & $R_4$$\downarrow$ \\
\midrule
\textbf{ELF-B} &  &  &  \\
W=128 & 19.86 & 5.53 & \apbest{5.16} \\
W=512 & 18.99 & 5.50 & 5.59 \\
G=8 & 19.01 & 5.50 & 5.64 \\
G=32 & 19.27 & 5.53 & 5.48 \\
K=4 & 20.54 & 5.67 & 5.28 \\
K=16 & \apbest{17.31} & 5.24 & 7.23 \\
early=0.115 & \apsecond{18.94} & 5.59 & 6.01 \\
early=0.250 & 19.54 & 5.50 & \apsecond{5.23} \\
\bottomrule[1pt]
\end{tabular}}
\end{adjustbox}
\apnote{Within model at fixed length: PPL $\downarrow$, $R_4\downarrow$; pending and unmatched rows unranked.}

\caption{\textbf{Hyperparameter sensitivity of ELF-B at 4k.} $W$, $G$, $K$ and the window-switch threshold are varied one at a time.}
\label{tab:hypersensitivity}
\end{table}

%% file: generated/table_enhanced_swa_keys.tex
\begin{table*}[!tp]
\centering\small
\begin{adjustbox}{max width=\linewidth}
{\lcstyle\begin{tabular}{ll|ccc|ccc|ccc}
\toprule[1pt]
 &  & \multicolumn{3}{c|}{B 8k} & \multicolumn{3}{c|}{M 8k} & \multicolumn{3}{c}{L 8k} \\
\cmidrule(lr){3-5}\cmidrule(lr){6-8}\cmidrule(lr){9-11}
Rung & Variant & PPL$\downarrow$ & $H$ & $R_4$$\downarrow$ & PPL$\downarrow$ & $H$ & $R_4$$\downarrow$ & PPL$\downarrow$ & $H$ & $R_4$$\downarrow$ \\
\midrule
Ref. & SWA(256,4) & 153.55 & 5.56 & 2.81 & 165.93 & 6.50 & 7.68 & 118.31 & 6.46 & 7.77 \\
Ref. & RangeSWA (Ours) & 18.37 & 5.63 & 6.24 & 32.13 & 6.21 & 1.23 & 33.39 & 6.36 & 0.79 \\
\midrule
B0 & Key-count-matched window & 162.11 & 5.68 & 4.16 & 133.51 & 6.40 & 1.87 & 92.97 & 6.42 & 3.07 \\
B1 & Key-count-matched window + corrections$^\dagger$ & 24.01 & 6.05 & 4.46 & 27.34 & 6.26 & 2.08 & 29.85 & 6.38 & 1.66 \\
\bottomrule[1pt]
\end{tabular}}
\end{adjustbox}
\apnote{Diagnostic values; no best/second ranking.}

\caption{\textbf{Sliding windows with a matched attention budget at 8k.} B0 uses a 1280-token window; B1 adds the conditioning corrections and the bounded decoder.}
\label{tab:enhancedswakeys}
\end{table*}

%% file: generated/table_enhanced_swa.tex
\begin{table*}[!tp]
\centering\small
\begin{adjustbox}{max width=\linewidth}
{\lcstyle\begin{tabular}{ll|ccc|ccc|ccc|ccc|ccc}
\toprule[1pt]
 &  & \multicolumn{3}{c|}{M 2k} & \multicolumn{3}{c|}{L 2k} & \multicolumn{3}{c|}{B 8k} & \multicolumn{3}{c|}{M 8k} & \multicolumn{3}{c}{L 8k} \\
\cmidrule(lr){3-5}\cmidrule(lr){6-8}\cmidrule(lr){9-11}\cmidrule(lr){12-14}\cmidrule(lr){15-17}
Rung & Variant & PPL$\downarrow$ & $H$ & $R_4$$\downarrow$ & PPL$\downarrow$ & $H$ & $R_4$$\downarrow$ & PPL$\downarrow$ & $H$ & $R_4$$\downarrow$ & PPL$\downarrow$ & $H$ & $R_4$$\downarrow$ & PPL$\downarrow$ & $H$ & $R_4$$\downarrow$ \\
\midrule
Ref. & SWA(256,4) & 29.40 & 5.78 & 6.94 & 29.09 & 5.69 & 5.55 & 153.55 & 5.56 & 2.81 & 165.93 & 6.50 & 7.68 & 118.31 & 6.46 & 7.77 \\
Ref. & RangeSWA (Ours) & 25.55 & 5.56 & 1.79 & 27.27 & 5.68 & 1.16 & 18.37 & 5.63 & 6.24 & 32.13 & 6.21 & 1.23 & 33.39 & 6.36 & 0.79 \\
\midrule
A0 & Sliding window, same code path & 29.71 & 5.78 & 6.86 & 29.42 & 5.71 & 5.35 & 171.70 & 5.56 & 2.56 & 158.71 & 6.52 & 7.81 & 121.22 & 6.47 & 7.39 \\
A1 & Sliding window + conditioning corrections$^\dagger$ & 21.94 & 5.65 & 7.38 & 28.68 & 5.62 & 3.60 & 120.42 & 6.34 & 2.50 & 42.87 & 6.31 & 5.19 & 32.96 & 6.23 & 2.57 \\
A2 & Sliding window + stage window schedule$^\dagger$ & 32.31 & 5.77 & 3.25 & 27.83 & 5.74 & 3.39 & 184.34 & 6.16 & 2.51 & 223.66 & 6.63 & 2.25 & 125.31 & 6.58 & 2.75 \\
A3 & Enhanced sliding window (both)$^\dagger$ & 22.96 & 5.66 & 3.97 & 27.29 & 5.73 & 2.17 & 28.95 & 6.20 & 4.64 & 24.31 & 6.28 & 3.60 & 28.89 & 6.36 & 1.94 \\
\bottomrule[1pt]
\end{tabular}}
\end{adjustbox}
\apnote{Diagnostic values; no best/second ranking.}

\caption{\textbf{Simplified variants of \method{} without remote information.} A0--A3 add the conditioning corrections and the window schedule to a plain sliding window.}
\label{tab:enhancedswa}
\end{table*}

%% file: generated/table_kernel_v2_profile.tex
\begin{adjustbox}{max width=\linewidth}
{\lcstyle\begin{tabular}{ll|rr|rrr}
\toprule[1pt]
\multicolumn{2}{c|}{} & \multicolumn{2}{c|}{Full} & \multicolumn{3}{c}{\method{}} \\
\cmidrule(lr){3-4}\cmidrule(lr){5-7}
Precision & Length & Attn. & Rest & Attn. & Prep. & Rest \\
\midrule
BF16 & 2k & \apbest{9.4} & 52.0 & 14.2 & 7.4 & \apbest{43.1} \\
 & 4k & 33.2 & 101.3 & \apbest{30.6} & 13.4 & \apbest{82.7} \\
 & 8k & 126.4 & 200.4 & \apbest{72.7} & 27.5 & \apbest{162.4} \\
\midrule
FP32 & 2k & \apbest{75.3} & 150.2 & 143.8 & 7.5 & \apbest{140.5} \\
 & 4k & \apbest{300.9} & 313.1 & 315.8 & 14.9 & \apbest{278.3} \\
 & 8k & 1155.3 & 608.7 & \apbest{767.4} & 29.8 & \apbest{545.9} \\
\bottomrule[1pt]
\end{tabular}}
\end{adjustbox}
\apnote{ms$\downarrow$; bold marks the lower of Full and \method{}, separately for attention and for the rest. Preparation unranked.}

%% file: generated/table_kernel_v2.tex
{\lcstyle\begin{tabular}{l|cccccc}
\toprule[1pt]
 & \multicolumn{3}{c}{BF16} & \multicolumn{3}{c}{Strict FP32} \\
\cmidrule(lr){2-4}\cmidrule(lr){5-7}
Method & 2k & 4k & 8k & 2k & 4k & 8k \\
\midrule
Full & 2.08 & 4.48 & 10.84 & 7.74 & 20.17 & 58.30 \\
\method{} Triton (Ours) & 2.36 (0.88$\times$) & 4.46 (1.01$\times$) & 8.98 (1.21$\times$) & 10.92 (0.71$\times$) & 23.42 (0.86$\times$) & 50.52 (1.15$\times$) \\
\lite{} Triton (Ours) & 1.81 (1.15$\times$) & 3.41 (1.32$\times$) & 6.71 (1.62$\times$) & 6.63 (1.17$\times$) & 13.41 (1.51$\times$) & 26.74 (2.18$\times$) \\
\bottomrule[1pt]
\end{tabular}}

%% file: generated/table_triton2_quality.tex
\begin{table*}[!tp]
\centering\small\setlength{\tabcolsep}{4pt}
\begin{adjustbox}{max width=\linewidth}
{\lcstyle\begin{tabular}{l|ccc|ccc|ccc|c}
\toprule[1pt]
 & \multicolumn{3}{c|}{2k} & \multicolumn{3}{c|}{4k} & \multicolumn{3}{c|}{8k} &  \\
\cmidrule(lr){2-4}\cmidrule(lr){5-7}\cmidrule(lr){8-10}
Triton kernel & PPL ($\Delta$) & $H$ & $R_4$$\downarrow$ & PPL ($\Delta$) & $H$ & $R_4$$\downarrow$ & PPL ($\Delta$) & $H$ & $R_4$$\downarrow$ & Worst layer err. \\
\midrule
\method{}, FP32 & \apbest{21.57 (-0.01)} & 5.36 & \apbest{4.6} & \apbest{19.03 (-0.03)} & 5.51 & \apbest{5.7} & \apbest{18.39 (+0.02)} & 5.63 & \apbest{6.2} & $1.71\times10^{-6}$ \\
\method{}, BF16 & \apbest{22.09 (+0.01)} & 5.37 & \apbest{4.3} & \apbest{19.41 (+0.03)} & 5.52 & \apbest{5.4} & \apbest{18.72 (+0.02)} & 5.65 & \apbest{5.9} & $1.71\times10^{-6}$ \\
\lite{}, FP32 & \apsecond{22.92 (+0.01)} & 5.58 & \apsecond{7.0} & \apsecond{21.50 (+0.02)} & 5.85 & \apsecond{8.0} & \apsecond{21.20 (-0.00)} & 6.09 & \apsecond{7.9} & $1.07\times10^{-6}$ \\
\lite{}, BF16 & \apsecond{23.15 (-0.07)} & 5.58 & \apsecond{6.9} & \apsecond{21.68 (-0.02)} & 5.86 & \apsecond{7.8} & \apsecond{21.43 (-0.05)} & 6.09 & \apsecond{7.7} & $1.07\times10^{-6}$ \\
\bottomrule[1pt]
\end{tabular}}
\end{adjustbox}
\apnote{Within length and precision: PPL $\downarrow$, $R_4\downarrow$.}

\caption{\textbf{Quality and numerical error of the Triton implementation.} ELF-B, 1000 samples per configuration; the last column is the maximum per-layer relative error.}
\label{tab:tritonquality2}
\end{table*}

%% file: generated/table_llm_judge.tex
\begin{table*}[!tp]
\centering\small
\begin{adjustbox}{max width=\linewidth}
{\lcstyle\begin{tabular}{ll|cl|cl|rrrrl}
\toprule[1pt]
\multicolumn{2}{c|}{} & \multicolumn{2}{c|}{Pairwise vs.\ SWA} & \multicolumn{2}{c|}{Pairwise vs.\ Full} & \multicolumn{5}{c}{Single-document score$\uparrow$} \\
\cmidrule(lr){3-4}\cmidrule(lr){5-6}\cmidrule(lr){7-11}
Model & Length & W/T/L & Net 95\% CI & W/T/L & Net 95\% CI & \method{} & SWA & Full & $\Delta$ & 95\% CI \\
\midrule
ELF-B & 2k & 65.0 / 33.0 / 2.0 & [+55.5, +70.0] & 52.5 / 37.0 / 10.5 & [+32.5, +51.5] & \apsecond{2.46} & 1.98 & \apbest{2.59} & -0.14 & [-0.23, -0.03] \\
 & 4k & 100.0 / 0.0 / 0.0 & --- & 100.0 / 0.0 / 0.0 & --- & \apbest{2.33} & \apsecond{1.54} & 1.00 & +1.33 & [+1.26, +1.40] \\
 & 8k & 100.0 / 0.0 / 0.0 & --- & 100.0 / 0.0 / 0.0 & --- & \apbest{2.54} & \apsecond{1.00} & \apsecond{1.00} & +1.54 & [+1.47, +1.61] \\
\midrule
ELF-M & 2k & 74.5 / 24.0 / 1.5 & [+66.5, +79.5] & 85.0 / 13.0 / 2.0 & [+76.5, +89.0] & \apsecond{2.19} & 2.00 & \apbest{2.20} & -0.01 & [-0.08, +0.07] \\
 & 4k & 98.5 / 1.5 / 0.0 & [+96.5, +100.0] & 100.0 / 0.0 / 0.0 & --- & \apbest{2.06} & \apsecond{1.44} & 1.01 & +1.05 & [+1.01, +1.09] \\
 & 8k & 100.0 / 0.0 / 0.0 & --- & 100.0 / 0.0 / 0.0 & --- & \apbest{2.03} & \apsecond{1.17} & 1.00 & +1.03 & [+1.01, +1.05] \\
\midrule
ELF-L & 2k & 94.0 / 6.0 / 0.0 & [+90.5, +97.0] & 83.0 / 13.0 / 4.0 & [+72.0, +85.5] & \apsecond{2.23} & 2.00 & \apbest{2.31} & -0.08 & [-0.17, +0.01] \\
 & 4k & 100.0 / 0.0 / 0.0 & --- & 100.0 / 0.0 / 0.0 & --- & \apbest{2.06} & \apsecond{1.84} & 1.00 & +1.05 & [+1.02, +1.09] \\
 & 8k & 100.0 / 0.0 / 0.0 & --- & 100.0 / 0.0 / 0.0 & --- & \apbest{2.03} & \apsecond{1.30} & 1.00 & +1.03 & [+1.01, +1.05] \\
\bottomrule[1pt]
\end{tabular}}
\end{adjustbox}
\apnote{W/T/L: \method{} wins/ties/losses in percent of 200 pairs; Net is W$-$L, and --- marks an interval degenerate at $+100$ because every pair was won. Single-document scores (1--5) ranked within each row; $\Delta$: \method{}$-$Full. Pairwise counts and contrasts unranked. Proxy ratings only.}

\caption{\textbf{Qwen3-14B evaluation of long-text coherence.} Pairwise win/tie/loss rates of \method{} against SWA and Full over 200 pairs, and single-document scores on a 1--5 scale.}
\label{tab:llmjudge}
\end{table*}